\pdfoutput=1
\documentclass[11pt]{article}
\usepackage[utf8]{inputenc}

\usepackage{acl}
\usepackage{times}
\usepackage{latexsym}
\usepackage{graphicx}
\usepackage{array}
\usepackage[table]{xcolor}
\usepackage{booktabs}
\usepackage{adjustbox} 
\usepackage{tabularx}
\usepackage{graphicx} 
\usepackage{array} 
\usepackage{setspace}
\usepackage{geometry}
\usepackage{longtable}
\usepackage{array}
\usepackage{caption}
\usepackage{lipsum}
\usepackage{fancyvrb}
\usepackage{multicol}
\usepackage{graphicx}
\usepackage{float}
\usepackage{adjustbox}
\usepackage{verbatim}

\usepackage[T1]{fontenc}
\usepackage[utf8]{inputenc}

\usepackage{microtype}

\usepackage{inconsolata}

\usepackage{tikz,lipsum,lmodern}
\usepackage[most]{tcolorbox}

\usepackage{pgfplots, pgfplotstable}
\pgfplotsset{compat=1.18}
\usepackage{lipsum} % (optional) to create filler text 
\usepackage{geometry}
\title{Does Reasoning Help LLM Agents Play Dungeons and Dragons? A Prompt Engineering Experiment}

\author{Patricia Delafuente, Arya Honraopatil, and Lara J. Martin \\
        University of Maryland, Baltimore County \\
        \texttt{\{pstanton,aryahonrao,laramar\}@umbc.edu}}

\date{}

\begin{document}

\maketitle

\begin{abstract}

This paper explores the application of Large Language Models (LLMs)  and reasoning to predict Dungeons \& Dragons (DnD) player actions and format them as Avrae Discord bot commands. Using the FIREBALL dataset, we evaluated a reasoning model, DeepSeek-R1-Distill-LLaMA-8B, and an instruct model, LLaMA-3.1-8B-Instruct, for command generation. Our findings highlight the importance of providing specific instructions to models, that even single sentence changes in prompts can greatly affect the output of models, and that instruct models are sufficient for this task compared to reasoning models.
\end{abstract}

\section{Introduction}
\label{sec:intro}
Dungeons \& Dragons (DnD) is a cooperative role-playing game where players create unique characters that embark on exciting adventures in a fantasy world. One player takes on the role of the Dungeon Master (DM), a special role to narrate the game, guide the story, and control non-player characters. The players interact with the DM and each other using natural language, often describing their intended \textit{actions}. The open-ended gameplay of DnD makes it an interesting and rich environment to test and challenge large language models (LLMs) \cite{martin2018dungeons,callison-burch-etal-2022-dungeons}.

LLMs have shown to help with text generation within the space of DnD, either helping players with the game commands and narration \cite{callison-burch-etal-2022-dungeons}, game design \cite{xiaozhan2024roleplayinggamedesign} or helping solve game tasks \cite{song2024thirteenhourssolvelabyrinth}. 
For an AI system to play DnD effectively, it needs strong language understanding, along with the ability to learn and represent knowledge, track game states, and reason over decisions \cite{callison-burch-etal-2022-dungeons}.
Thanks to reasoning mechanics, LLMs have recently undergone a tremendous transformation in their capacity to solve more complex problems.
\textit{Reasoning LLMs} such as Deepseek R1 \cite{deepseekai2025deepseekr1incentivizingreasoningcapability} are finetuned models that perform chain-of-thought (CoT) reasoning \cite{wei2022chain} iteratively.
These models have evolved LLMs from token-level prediction to systems capable of multi-step thought---improving performance in arithmetic, logic and common sense reasoning tasks \cite{shao2024deepseekmathpushinglimitsmathematical}.

In this paper, we explore the impact of prompt design on reasoning and instruction-tuned LLMs for predicting the generation of structured actions for players in the game of DnD. This work focuses on the generation of actions during combat, such as an attack on a non-player character or monster using an available weapon in the player's inventory. This work is a stepping stone for future work on using LLMs to predict a recommended action for the current player at any point in the game.

Specifically, we look at the capabilities of both types of LLMs for generating commands for the Avrae Discord bot. Avrae is a bot that enables players to play DnD virtually by allowing them to use commands to perform in-game actions, like rolling dice or casting spells, all while tracking the state of entities in the game world.
The LLM is instructed to generate the recommended attack and applicable attributes such as the weapon and target as a structured Avrae command. 
In Avrae, if, for example, \textit{Fangling} is a character who wants to attack a target with one of their available weapons such as a \textit{2-Handed Longsword}, an expected Avrae command would be \texttt{!a 2-Hand}. 

Avrae has been used in the past to look at how well LLMs can translate natural language narration to the structured representation needed to run Avrae and vice versa \cite{papazov2022avrae,Zhu_2023}. We are extending this work by exploring whether this can be considered a ``reasoning'' problem appropriate for reasoning LLMs.
This paper addresses the following questions:
\begin{enumerate}
\item Is reasoning an effective technique for generating game actions given a game state?
\item What types of prompting techniques make reasoning or instruct models better or worse for this task?
\end{enumerate}

Generating gameplay commands using LLMs helps models learn structured natural language grounding, intent recognition, and world-state reasoning. These skills can advance interactive storytelling by enabling more coherent, context-aware, and character-consistent narrative systems and game automation.

\section{Related Work}

\subsection{Reasoning Models}
Early models used token prediction for knowledge retrieval but were limited in their ability to solve complex tasks \cite{rae2021scaling}.

The transition towards reasoning began with Chain-of-Thought (CoT) prompting \cite{wei2022chain}, which showed that LLMs could generate more accurate solutions by prompting the model to "think step by step". Building on this,  \citet{wang-etal-2023-self-instruct}, showed that sampling multiple reasoning paths and aggregating them yielded more reliable answers. 

Other approaches like Least-to-Most prompting \cite{zhou2023leasttomost} re-framed complex problems into simpler sub-problems, solved sequentially, and methods such as Tree-of-Thoughts \cite{yao2023tree} and Graph-of-Thoughts \cite{maciej2024graph} formalize reasoning as an explicit exploration over possible inference paths. 

Modern reasoning models like OpenAI's o1 \& o3 and DeepSeek's DeepSeek-R1 incorporate reinforcement learning and adaptive computation to allow for additional "thinking time" for complex queries \cite{jaech2024openai, deepseekai2025deepseekr1incentivizingreasoningcapability}. Reasoning models extend the CoT paradigm by fine-tuning the models so that they generate their reasoning automatically without being explicitly prompted to do so. 

Although reasoning models have shown improvements in performance on some tasks \cite{illusion-of-thinking}, it is still unknown what types of tasks they are well suited for, which inspires this work.
As the complexity of the problem increases, the reasoning effort declines past a certain point and the performance of the model falls \cite{illusion-of-thinking}.

\subsection{LLMs for DnD}
LLMs have played a crucial role in the development of interactive agents and text-based games \cite{shuster-etal-2022-language, ammanabrolu2021knowledge, urbanek2019learning}. 
Early work on DnD by \citet{Zhu_2023} showed that LLMs, such as GPT-3, can map natural language play descriptions to executable Avrae commands with moderate accuracy. 
We build on the work of \cite{Zhu_2023} to demonstrate if reasoning models are required to automate command generation or if decoder models are sufficient for the task. 
Others have used LLMs to generate DnD components, such as spells \cite{musacchio-etal-2024-leveraging}. 

In their work, \citet{musacchio-etal-2024-leveraging} state that LLMs (such as GPT-2, OPT, LLaMA-2) were able to produce diverse and imaginative spell descriptions but often struggled with consistency, balance, and mechanical validity, thus underscoring the need for reasoning-oriented approaches for structured rule constraints. 

Similarly to \citet{callison-burch-etal-2022-dungeons}, we use game state information to add context to the LLM prompt. However, instead of generating dialog, we are generating a character's next action.

\section{Experimental Setup}

We look at the task of generating game action commands for the DnD Discord bot Avrae.
To answer our research questions (\S\ref{sec:intro}), we compare the output of two models -- an instruct model and a reasoning model, both fine-tuned from the same original model (LLaMA-3.1-8B).

To compare the two models, we performed a prompt engineering experiment with qualitative analysis and then compared the generated commands across the models and prompts using quantitative results.

We explain the prompt engineering experiment in Section \ref{sec:prompting} and the quantitative measures in Section \ref{sec:quant}.

\subsection{Data}
Our goal is to evaluate the accuracy and quality of command predictions generated, comparing a reasoning and instruct model across different system prompt variations. To do this, we need a dataset of DnD game plays with sufficient game state on the current and other players that can be used as input to enable LLMs to generate an action command for the current player.

We used the FIREBALL dataset \cite{Zhu_2023}, available on HuggingFace\footnote{\url{https://huggingface.co/datasets/lara-martin/FIREBALL}}. This dataset contains more than 100,000 instances of DnD in-game actions and game state scraped from DnD games played on Discord. The game actions are formatted as commands that are executed for the Avrae Discord bot. Avrae is a bot that helps people play text-based DnD games virtually on Discord. It allows players to interact with each other, simulate dice rolls, and execute commands that represent a specific character action. The dataset contains information such as gameplay utterances and commands, the current combat state, the current player, and any updates to the state after an action is taken. Each action has a paired Avrae command.  
\begin{singlespace}
We used the following information from FIREBALL: 
\begin{itemize}
    \item \textbf{current\_actor}: Game state information for the current player. 
    \item \textbf{combat\_state\_before}: Game state for all players in the game.
    \item \textbf{utterance\_history} In-character dialogue of the players for the current and previous turns.  
    \item \textbf{commands\_norm}: Avrae command for the current action.  This is the expected output of our models.
\end{itemize}
\end{singlespace}
\noindent
An abbreviated combat\_state\_before example is shown below, but the full example is found in the appendix.

\vspace{0.5em}

\begin{small}
\begin{tcolorbox}[breakable,colback=red!3!white,colframe=red!75!black,title=\textbf{current\_actor}]
\texttt{{\textcolor{red}{"name"}: "Emma Thornwall", \\ 
\textcolor{red}{"hp"}: "<80/80 HP; Healthy>", \\ 
\textcolor{red}{"class"}: "Warlock 11", \\ 
\textcolor{red}{"race"}: "Eladrin", \\ 
\textcolor{red}{"attacks"}: "Crossbow, light, Dagger, Quarterstaff, 2-Handed Quarterstaff, Scross, Silver, Unarmed Strike", \\ 
\textcolor{red}{"spells"}: "Levitate, Magic Missile, Hold Person, Counterspell, Witch Bolt, Intellect Fortress, Dimension Door, Raulothim's Psychic Lance, Sending, Death Ward, Polymorph, Hex, Mirror Image, Eyebite, Eldritch Blast, Prestidigitation, Mage Hand, Dissonant Whispers, Mind Sliver", \\ 
\textcolor{red}{"actions"}: "Sculptor of Flesh, Fey Step (Winter), Protection of the Talisman, Fey Step (Summer), Fey Step (Spring), Agonizing Blast, Pact of the Talisman, Fey Step (Autumn), Entropic Ward, Ascendant Step", \\ 
\textcolor{red}{"effects"}: "Blessed", \\ 
\textcolor{red}{"description"}: null, \\ 
\textcolor{red}{"controller\_id"}: "178818952437053304"}}\\
\end{tcolorbox}

\begin{tcolorbox}[breakable,colback=red!3!white,colframe=red!75!black,title=\textbf{combat\_state\_before} -- each listed character contains all of the information found in current\_actor]
\texttt{\textcolor{red}{"name"}: "Emma Thornwall", \\ 
..., \\
\textcolor{red}{"name"}: "BA1", \\ 
..., \\ 
\textcolor{red}{"name"}: "Jaguar", \\ 
...,\\
\textcolor{red}{"name"}: "Zenthaea", \\ 
..., \\
\textcolor{red}{"name"}: "Petcan Gard", \\ 
...}
\end{tcolorbox}

\vspace{0.5em}

\begin{tcolorbox}[breakable,colback=red!3!white,colframe=red!75!black,title=\textbf{utterance\_history}]
\texttt{\textcolor{red}{"Player 2}: go ahead and recast it"}
\end{tcolorbox}

\vspace{0.5em}

\begin{tcolorbox}[breakable,colback=red!3!white,colframe=red!75!black,title=\textbf{commands\_norm}]
\texttt{"!cast psychic -t ba1"}
\end{tcolorbox}

\end{small}

\vspace{0.5em}

We pulled a sample of 4071 rows from the FIREBALL dataset. We processed the \texttt{current\_actor} and \texttt{combat\_state\_before} fields to remove attributes that are not needed, such as \texttt{controller\_id}. Rows that contained obvious anomalies, such as the missing data, and rows longer than 4001 characters were removed, as well. 
To create the input prompts, we merged the processed \texttt{current\_actor}, \texttt{combat\_state\_before}, and \texttt{utterance\_history} fields, prefixed the merged fields to the system prompt.  There were five variations of the system instruction prompt, so each unique game state row has five instances in the dataset for a total dataset row size of 20,355.

\subsection{Models}
We selected LLaMA-3.1-8B-Instruct and DeepSeek-R1-Distill-LLaMA-8B for our instruct model and reasoning model. 
We will refer to these as Instruct and R1-Distill, respectively. Note that both models are based on LLaMA-3.1 with 8 billion parameters.

R1-Distill is a reasoning model that was trained to handle tasks that may benefit from intensive reasoning or CoT. 
It is noted to excel in scientific and logic reasoning tasks that are well defined and have clear solutions \cite{deepseekai2025deepseekr1incentivizingreasoningcapability}. It shows strong performance in open-ended tasks, such as creative writing and general questions \cite{li2025reasoningtrustingbehaviordeepseek}, but the DeepSeek-R1 developers indicated that more exploration is needed in multi-turn or role-playing tasks.  The developers also recommend that users directly describe the problem and specify the output format using a zero shot scenario rather than a few shot prompts for optimal results \cite{deepseekai2025deepseekr1incentivizingreasoningcapability}.

The LLaMA family of instruction-following models, on the other hand, have extensive training on a wide range of conversational and narrative data and post-training alignment with human preferences that enable them to follow instructions, maintain a natural flow in role-playing contexts, and adapt to various personas, tones, and scenarios \cite{grattafiori2024llama3herdmodels}.

\subsection{Evaluation Metrics}
\label{sec:quant}
To evaluate the quality of the generated commands, we directly compared them to the ground-truth Avrae commands written by players in the FIREBALL dataset.   Specifically, we calculated BLEU \cite{bleu}, ROUGE-1/ROUGE-2/ROUGE-L \cite{rouge}, and perplexity.
However, since numerous actions and arguments (e.g., weapons, targets) can be generated while still creating a valid Avrae command, we did not expect the commands generated by the LLMs to match the ground-truth Avrae commands. Thus, we created custom metrics---Reference and Format Checks---to evaluate the quality and format of the Avrae command generated by the LLM. 
Specifically, they are:

\paragraph{Format Check.} Avrae has a specific syntax that it expects and will not process commands outside of this format. Therefore, all generated commands must conform to this. The format usually contains a command name and arguments, such as: 

\texttt{![command name] [arg0] [flag1] [arg1] [flag2] [arg2] ...}

\noindent
For example, the proper command format to initiate an attack is 

\texttt{!attack <attack name> -t <target name>}

\noindent
and to cast a spell is 

\texttt{!cast <spell name> -t <target name>}

\noindent
The Format Check is calculated as the percentage of generated commands across the dataset that are in the correct format.

\paragraph{Reference Check.} For the command to work, it needs to reference entities that have previously been mentioned in the game state. This enables Avrae to keep track of and update the game state and enemies \& players within it as the combat continues. 
The Reference Check compares the arguments of the generated command with the entities in the game state, verifying that the attack type or spell type are actions that the player can take or that the target of the attack is found in the game state. 
Any command that has an attack mismatch, spell mismatch, target mismatch, or no command was generated at all is considered to fail the Reference Check.
We then calculate the percentage of commands that have fully correct references.

We used all of these metrics to recognize patterns between the two models across the five prompt variations.

\begin{table*}[ht]
    \centering
    \begin{tabularx}{14cm}{c c c c c c c c c}
        \toprule

        \multicolumn{9}{l}{\textsc{LLaMa-3.1-8B-Instruct}} \\       
        
               \textbf{Prompt} & \textbf{Format}$\uparrow$ 
               & \textbf{Perplexity} $\downarrow$ & \textbf{Rge-1}$\uparrow$ & \textbf{Rge-2}$\uparrow$ & \textbf{Rge-L}$\uparrow$ & \textbf{BLEU}$\uparrow$ \\
                \hline
         Prompt1 & 0.988 & \textbf{1.157} & 0.229 & 0.081 & 0.223 & 0.0017 \\
         Prompt2 & \underline{0.994}  & 1.189 & \underline{0.253} & \underline{0.087} & 0.249 & 0.0025 \\
         Prompt3 & \textbf{0.995} & 1.228 & \underline{0.253} & 0.086 & \underline{0.250} & 0.0023 \\
         Prompt4 & 0.987 & \textbf{1.157} & 0.229 & 0.086 & 0.248 & 0.0026 \\
         Prompt5 & 0.981 &  1.189 & 0.221 & 0.077 & 0.216 & \underline{0.0030} \\
        \midrule
        \multicolumn{9}{l}{\textsc{DeepSeek-R1-Distill-LLaMa-8B}} \\
        
        \textbf{Prompt} & \textbf{Format}$\uparrow$   &  \textbf{Perplexity} $\downarrow$ & \textbf{Rge-1}$\uparrow$ & \textbf{Rge-2}$\uparrow$ & \textbf{Rge-L}$\uparrow$ & \textbf{BLEU}$\uparrow$ \\
        \hline
         Prompt1 & 0.992   &  1.189 & \textbf{0.256} & \textbf{0.089} & \underline{0.250} & 0.0022 \\
         Prompt2 & 0.992  & \underline{1.184} & 0.250 & 0.086 & 0.245 &  \underline{0.0030} \\
         Prompt3 & \underline{0.994}   & 1.189 & 0.215 & 0.073 & 0.210 & 0.0026 \\
         Prompt4 & 0.992 &1.188 & 0.222 & 0.074 & 0.212 & 0.0021 \\
         Prompt5 & 0.992 & 1.188 & \textbf{0.256} & \textbf{0.089} & \textbf{0.251} & \textbf{0.0036} \\
        \bottomrule
    \end{tabularx}
    \caption{Summary of evaluation metrics comparing an instruct and reasoning model across five prompts. Bolded values indicate the best prompt/model combination. Underlined values are second best. Reference Check scores are found separately in Figure \ref{fig:referencecheck}. Rge-1/2/L correspond to ROUGE-1, ROUGE-2, ROUGE-L.}
    \label{tab:model_summary}
\end{table*}

\section{Prompting Experiments}
\label{sec:prompting}

Typically, the prompt engineering process is excluded from scientific publications. However, we find that, like all design processes, the iterative process of prompt engineering tells an interesting story. 

For the prompt engineering experiment, we began with an initial prompt (Prompt 1) and ran it over the dataset to generate commands. The prompt was identical for both models. The researcher (the first author) then looked over the generated output and made a judgment, based (quantitatively and qualitatively) on the quality of the output, of what changes to make to the prompt. The procedure repeated on the subsequent prompt (Prompt 2), and the prompts were continued until the researcher was content with the output of the model. In this study, we ended up with five prompts.

In this section, we walk through each of the prompts and our reasoning behind the changes. Changes from one prompt to another are highlighted in red text.
All five prompts began with the following paragraph:
\begin{tcolorbox}[breakable,colback=blue!5!white,colframe=blue!75!black,title=Beginning Paragraph of all Prompts]
You are the current player in a Dungeons and Dragons game played in Discord. An Avrae Bot is used to enable players to type in commands to roll dice and initiate actions in Discord. Your job is to determine an action for the current player and then construct a properly formatted Avrae command to initiate that action. You will receive as input, the current game state info of the current player, game state of the other players and history of the in game utterances. Output the final Avrae command within \{\{\}\}.
\end{tcolorbox}

\subsection{Prompt 1: Attack Example}
We began our prompting experiment by giving Prompt 1 format guidance and one example of an Avrae command:

\texttt{!attack longbow -t WY1 adv}

\noindent
We also told the model to reason step-by-step, which should force CoT reasoning.

\begin{tcolorbox}[breakable,colback=blue!5!white,colframe=blue!75!black,title=Prompt 1]
If the action is an attack, the format of an Avrae for an attack is \texttt{!attack <attack name> -t <target name> [arguments]} where you replace <attack name> with the desired attack and <target> with the name of the target and then a list of any applicable arguments. \\ 
For example the command, `\texttt{!attack longbow -t WY1 adv}' would initiate an attack using the longbow weapon in the player inventory against the target WY1 with an advantage argument. \\ 
Reason step by step, and then provide your final answer. 
\end{tcolorbox}

We observed in the reasoning trace that the R1-Distill CoT placed an unnecessary emphasis on the weapon in the example and selected an attack when one might cast a spell. 
For example, post-inference dataset row 1349 has player data:
\begin{quote}
\footnotesize
\texttt{Current Player: \\ {`name': Cressida Caldwell \\ `class': Witch 6 \\`hp': <8/38 HP; Bloodied> \\ `attacks': Unarmed Strike, Dagger \\ `spells': Ray of Sickness, Command, Healing Word, Enthrall, Lesser Restoration, Charm Person, Chill Touch, Guidance, Dissonant Whispers, Slow, Suggestion, Moonbeam, Hypnotic Pattern, Incite Greed, Toll the Dead, Find Familiar, Fast Friends, Bane}}
\end{quote}

Due to the class of the player and the wealth of spells, one might opt to cast a spell rather than an attack. The ground truth Avrae command selected by the player is \texttt{!cast diss -t zo2}, indicating they chose an action to cast a spell.  However, the reasoning trace shows how the model is influenced by the attack example used in the prompt: 
\begin{quote}
\textit{ [...] wait, but the example given uses a longbow against WY1 with an advantage. So in this case, Cressida's Dagger is a melee attack, so she can use that against a target. So the target would be Hylga, as she's injured. The attack name is Dagger, and the target is Hylga. So the command would be !attack Dagger -t Hylga. Alternatively, if she uses a spell, it would be !spell Ray of Sickness -t Hylga. But the user's example used an attack, so maybe she wants to attack with Dagger. [...]}
\end{quote}
R1-Distill then finished its reasoning by generating the command
\texttt{!attack Dagger -t Hylga}.

The Instruct model was not influenced by the example and predicted the command \texttt{!spell Healing Word -t Inquisitus}. Note that the generated command should have used \texttt{cast} in place of \texttt{spell}, so even the Instruct model has some issues with proper command generation with Prompt 1. 

Format Checks with Prompt 1 show that both models are able to properly format the Avrae command scoring 99\%.  For Reference Checks---to confirm that valid attacks, spells, and targets were selected---Instruct scored 55\% and R1-Distill scored 45\%.  Instruct outperformed R1-Distill, but the scores indicated that improvement was needed.

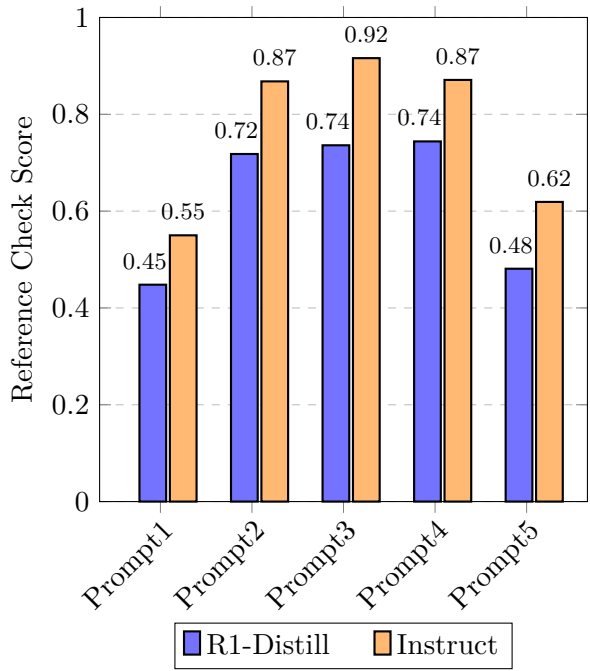
\begin{figure}[ht]
\centering
\begin{tikzpicture}

\pgfplotsset{
  /pgfplots/bar legend/.style={
    /pgfplots/legend image code/.code={
      \draw[##1,/tikz/.cd,bar width=6pt,yshift=-0.3em]
        plot coordinates {(0cm,0.8em)};
    },
  },
}

\begin{axis}[
    ybar,
    bar width=10pt,
    width=8cm,
    height=8cm,
    ymin=0, ymax=1.0,
    ylabel={Reference Check Score},
    symbolic x coords={Prompt1,Prompt2,Prompt3,Prompt4,Prompt5},
    xtick=data,
    xticklabel style={xshift=-8pt,rotate=45,yshift=5pt},
    nodes near coords,
    every node near coord/.append style={font=\small, text=black},
    nodes near coords align={vertical},
    legend style={at={(0.5,-0.25)}, anchor=north, legend columns=-1},
    enlarge x limits={abs=0.8cm},
    ymajorgrids=true,
    grid style=dashed,
    clip=false,
    legend style={/tikz/every even column/.append style={column sep=0.5cm}},
    bar legend,  
]

\addplot+[ybar,
          bar shift=-0.1cm,
          fill=blue!55!white,
          draw=black,
          thick,
          nodes near coords,
          nodes near coords style={yshift=2pt, xshift=-3pt, anchor=south, font=\small, text=black}
         ]
coordinates {
    (Prompt1,0.448)
    (Prompt2,0.718)
    (Prompt3,0.736)
    (Prompt4,0.744)
    (Prompt5,0.481)
};

\addplot+[ybar,
          bar shift=0.30cm,
          fill=orange!55!white,
          draw=black,
          thick,
          nodes near coords,
          nodes near coords style={yshift=2pt, anchor=south, font=\small, text=black}
         ]
coordinates {
    (Prompt1,0.550)
    (Prompt2,0.868)
    (Prompt3,0.916)
    (Prompt4,0.871)
    (Prompt5,0.619)
};

\legend{R1-Distill, Instruct}
\end{axis}
\end{tikzpicture}
\caption{Reference Check percentage across all prompts for both models. Higher is better.}
\label{fig:referencecheck}
\end{figure}

\subsection{Prompt 2: Attack and Cast Example, Do not refer to examples}

For Prompt 2, we decided to use both an attack and a spell example with additional guidance for each, such as checking that the attack or spell is one that exists in the player inventory and the target is one of the other players or non-player characters. 
This prompt also included specific guidance that the examples are not relevant to the current game state and to reason step by step.  

\begin{tcolorbox}[breakable,colback=blue!5!white,colframe=blue!75!black,,title=Prompt 2]
If the action is an attack, the format of the Avrae command for an attack is \texttt{!attack <attack name> -t <target name> [arguments]} where you replace <attack name> with the desired attack \textcolor{red}{and weapon that is available in the players inventory} and <target> with the name of the target \textcolor{red}{which can be one of the other players} and then a list of any applicable arguments. \\ 
\textcolor{red}{If the action is a cast then the format is \texttt{!cast <spell name> -t <target name> [arguments]} where you replace spell with a spell in the players inventory and <target name> with one of the other players and then a list of any applicable arguments.} \\ 
\textcolor{red}{Here are two example commands are }\texttt{!attack longbow -t WY1 adv} \textcolor{red}{to} initiate an attack using the longbow weapon against WY1 with an advantage argument \textcolor{red}{and \texttt{!cast `fire bolt' -t BA3} which casts Fire Bolt at BA3. \\ 
These examples are not relevant to current game state.} \\
\noindent%
Reason step by step and provide your final answer. 
\end{tcolorbox}

Both models scored well on the Format of the generated Avrae command with results of 99\%.  We saw notable improvements in the quality of the Avrae commands with the Reference Check score for Instruct increasing to 87\% and R1-Distill increasing to 72\%, although Instruct still outperformed R1-Distill.

\subsection{Prompt 3: Attack and Cast Example, Do not refer to examples, No Explicit Reasoning}

For Prompt 3, we decided to modify Prompt 2 to exclude the instruction to reason step by step and provide your final answer.
Although Prompt 2 showed improvements in the accuracy of the generation of the Avrae command in both models, there is a question of how the explicit instruction to \textit{reason step by step} may affect have effected the generation.  

Prompt 3 is the same as Prompt 2 except that the sentence
``Reason step by step and provide your final answer.'' was removed from the end.
The idea here is to evaluate the importance of reasoning to this specific task of next action prediction and Avrae command generation given the current state of the game. We expect that the R1-Distill model may still reason, but to a lesser extent. 

\begin{tcolorbox}[breakable,colback=blue!5!white,colframe=blue!75!black,title=Prompt 3]
 If the action is an attack, the format of the Avrae command for an attack is \texttt{!attack <attack name> -t <target name> [arguments]} where you replace <attack name> with the desired attack and weapon that is available in the players inventory and <target> with the name of the target which can be one of the other players and then a list of any applicable arguments. \\
 If the action is a cast then the format is \texttt{!cast <spell name> -t <target name> [arguments]} where you replace spell with a spell in the players inventory and <target name> with one of the other players and then a list of any applicable arguments.\\ 
 Here are two example commands are \texttt{!attack longbow -t WY1 adv} and \texttt{!cast `fire bolt' -t BA3} which casts Fire Bolt at BA3. \\ 
 \textcolor{red}{These examples are not relevant to current game state.}
\end{tcolorbox}

Both models score well when checking the format of the generated Avrae command (Figure \ref{tab:model_summary}). 
Interestingly, the Reference Check score for the Instruct model increased from 87\% to 92\% and R1-Distill had a small increase from 72\% to 74\%.  
For smaller decoder models, CoT has been observed to decrease performance, particularly for those with 10B parameters or smaller \cite{yin2025enhancinggeneralizationchainthought}. We see this same trend with our Instruct model.

\subsection{Prompt 4: Zero-shot, Explicit Reasoning}

By now, we now have observed that providing more detailed instruction on the construction of the Avrae command can improve accuracy. For the Instruction model, it may be more efficient not to use reasoning, as shown by the increase in Reference Check scores for both models with Prompt 3  (See Figure \ref{fig:referencecheck}). 

We were curious what would happen if we did not provide examples at all. We decided to test this and took Prompts 1 and 2, kept the instruction but removed the examples to make them zero-shot prompts.   
Prompt 4 is the same as Prompt 2 except the sentences 
\begin{quote}
``Here are two example commands are \texttt{!attack longbow -t WY1 adv} and \texttt{!cast `fire bolt' -t BA3} which casts Fire Bolt at BA3. \\
These examples are not relevant to current game state.'' 
\end{quote}
\noindent
were removed. Like Prompt 2, the sentence
``Reason step by step and provide your final answer.'' was added to the end of the prompt.

\begin{tcolorbox}[breakable,colback=blue!5!white,colframe=blue!75!black,title=Prompt 4]
 If the action is an attack, the format of the Avrae command for an attack is \texttt{!attack <attack name> -t <target name> [arguments]} where you replace <attack name> with the desired attack and weapon that is available in the players inventory and <target> with the name of the target which can be one of the other players and then a list of any applicable arguments. \\ If the action is a cast then the format is \texttt{!cast <spell name> -t <target name> [arguments]} where you replace spell with a spell in the players inventory and <target name> with one of the other players and then a list of any applicable arguments. \\ \textcolor{red}{Reason step by step, and then provide your final answer.}
\end{tcolorbox}

Instruct had the same results as Prompt 2 with a reference score of 87\% and was less than its score for Prompt 3 which was 92\%.  R1-Distill's score for Prompt 4 was slightly increased from Prompt 2 and similar to the results of Prompt 3 with a score of 74\%.

Thus, it appears that providing in-depth structure guidance is more important than examples for reasoning models.   We also observed that the Instruct model benefited from in-depth structure guidance and that, even in the zero-shot setting where reasoning models are meant to thrive, it outperforms R1-Distill.

\subsection{Prompt 5: Zero-shot, Attack Format Only, Explicit Reasoning}
Finally, we end with the simplest prompt. Prompt 5, like Prompt 1, has no command examples, but differs in that it leaves out the format structure for casting spells.

\begin{tcolorbox}[breakable,colback=blue!5!white,colframe=blue!75!black,,title=Prompt 5]
If the action is an attack, the format of an Avrae for an attack is \texttt{!attack <attack name> -t <target name> [arguments]} where you replace <attack name> with the desired attack and <target> with the name of the target and then a list of any applicable arguments.  Reason step by step and then provide your final answer.
\end{tcolorbox}

The Reference Check scores dropped significantly compared to Prompts 2, 3, and 4 but performed better than Prompt 1 with a score of 62\% for Instruct and 48\% for R1-Distill. 

Interestingly, not including an attack example like Prompt 1 showed slightly improved Reference Check scores for both models when comparing Prompt 5 which to Prompt 1. The R1-Distill again predicted an attack command instead of a spell for line 1349: 
\texttt{!attack Aid -t self}.

\section{Overall Analysis}

The Reference Check was designed to confirm that selected attacks, spells, weapons, and targets are appropriate such that a selected attack or spell is one that exists in the players' inventory and that the target matches one of the other players. To see how Reference Check scores compared across models and prompts, refer to Figure  \ref{fig:referencecheck}. We found that the prompts matter quite a bit here. There were significant improvements for both models with Prompts 2, 3, and 4 compared to Prompts 1 and 5, which had simpler prompts with less command structure guidance.  
Across all prompts, Instruct excels in Reference Check accuracy, suggesting stronger reasoning and grounding in the game state compared to R1-Distill (see Table \ref{tab:model_summary}). The Instruct scores were  87\% for Prompt 2, 92\% for Prompt 3, and 87\% for Prompt 4,  while R1-Distill was 72\%, 74\%, and 74\%, respectively.

A summary of the other quantitative results can be viewed in Table \ref{tab:model_summary}.
Both models maintain excellent Format compliance with scores of at least 99\% or higher across all prompts on the Format Checks, showing strong syntactic control to understand the structure of an Avrae command.
Both models for the five prompts had low perplexity scores, indicating high confidence, suggesting that the generated text is likely to be coherent in terms of output structure. Instruct was slightly higher on Prompt 3 with a score of 1.228, but the overall difference was minimal.

We expected the ROUGE and BLEU scores to be low as the predicted commands are compared to the actual player commands, of which the player had a variety of actions to choose from. 
The highest ROUGE and BLEU scores came from the R1-Distill model, although all of the scores were very close to each other. Prompt 5 performed consistently the best across all ROUGE \& BLEU metrics.
The BLEU scores across all models for all prompts are low, indicating that the generated text has very little overlap with the ground-truth text in terms of n-grams, as expected.  The BLEU scores were lowest for Instruct with Prompt 1 at 0.0017 and for R1-Distill with Prompt 4 at 0.0021.

Overall, the Instruct performed best with Prompt 3 which supports previous research that CoT reasoning can degrade performance for smaller LLMs \cite{yin2025enhancinggeneralizationchainthought}.
The developers of DeepSeek-R1
\cite{deepseekai2025deepseekr1incentivizingreasoningcapability} recommended zero shot paired with sufficient structure guidance rather than few shot prompts for optimal results. Our results support this, but we also find that if examples are used, results can be improved if they are paired with sufficient instruction to avoid providing extra weight to those exact examples in reasoning. This was evident with an increase in Reference Check scores for Prompts 2 \& 3 compared to Prompts 1 \& 5. Prompts 2 \& 3 (with examples) were comparable to Prompt 4 which did not use examples.

\section{Discussion \& Conclusion}

In this paper we tried to answer the question---can reasoning help LLM agents play DnD? For the specific task of generating an action for the Avrae Discord bot given DnD game state, it can, but instruct models are sufficient and can outperform reasoning models. Additionally, when using smaller models, forcing CoT reasoning can degrade performance \cite{yin2025enhancinggeneralizationchainthought}. This study explored the use of LLMs, comparing an instruct and a reasoning model with a focus on qualitative analysis of the output across five different prompt variations.    

Instruct models, such as LLaMA-3.1-8B-Instruct, combined with effective prompt engineering, are sufficient to generate accurate and logically valid commands. Although reasoning models such as DeepSeek-R1-Distill-LLaMA-8B may excel in complex reasoning tasks, their computational cost as noted in the Compute Details section(\ref{sec:app-compute}) of the Appendix during inference does not yield significant advantages for the type of open-ended task used in our experiments.   

\begin{singlespace}
Based on our observations, we offer the following guidance on the task of using LLM agents to evaluate game state to select and format an action in an open-ended environment:  
\begin{itemize}
    \item \textbf{Avoid CoT}: Avoid forcing CoT reasoning for smaller decoder models. CoT has been observed to decrease performance, particularly those under 10B parameters or smaller \cite{yin2025enhancinggeneralizationchainthought} and may not have enough context for step-by-step reasoning. 
    \item \textbf{Format}: Provide detailed instructions on the syntax, rules, and calculations needed to construct commands \cite{deepseekai2025deepseekr1incentivizingreasoningcapability}. For reasoning models, either use zero-shot prompting \cite{deepseekai2025deepseekr1incentivizingreasoningcapability} or, if examples are used, add instructions to ignore them. This helps the model to not place unnecessary emphasis on examples that are not relevant to the current problem.   
\end{itemize}
\end{singlespace}

\section*{Limitations}
Due to computational constraints, we were restricted to smaller (8B parameter) model. 
Future work is needed to explore and compare results with larger instruct and reasoning models, other types of tasks, and incorporate generated commands LLM agents into automated gameplay. That said, smaller models that can fit onto a single GPU are more likely to be used and therefore their utility should be studied.

We were interested in what the models knew off-the-shelf and therefore did not finetune them. They would have most likely performed better if they were finetuned.

\bibliography{avraeDnD}

\begin{thebibliography}{25}
\providecommand{\natexlab}[1]{#1}

\bibitem[{Ammanabrolu and Riedl(2021)}]{ammanabrolu2021knowledge}
Prithviraj Ammanabrolu and Mark~O. Riedl. 2021.
\newblock \href {https://proceedings.neurips.cc/paper_files/paper/2021/hash/1e747ddbea997a1b933aaf58a7953c3c-Abstract.html} {{Learning Knowledge Graph-based World Models of Textual Environments}}.
\newblock In \emph{Proceedings of the 35th International Conference on Neural Information Processing Systems}, volume~34 of \emph{NeurIPS '21}, pages 3720--3731, Red Hook, NY, USA. Curran Associates Inc.

\bibitem[{Besta et~al.(2024)Besta, Blach, Kubicek, Gerstenberger, Podstawski, Gianinazzi, Gajda, Lehmann, Niewiadomski, Nyczyk, and Hoefler}]{maciej2024graph}
Maciej Besta, Nils Blach, Ales Kubicek, Robert Gerstenberger, Micha\l{} Podstawski, Lukas Gianinazzi, Joanna Gajda, Tomasz Lehmann, Hubert Niewiadomski, Piotr Nyczyk, and Torsten Hoefler. 2024.
\newblock \href {https://doi.org/10.1609/aaai.v38i16.29720} {{Graph of Thoughts: Solving Elaborate Problems with Large Language Models}}.
\newblock \emph{Proceedings of the Thirty-Eighth AAAI Conference on Artificial Intelligence and Thirty-Sixth Conference on Innovative Applications of Artificial Intelligence and Fourteenth Symposium on Educational Advances in Artificial Intelligence}, 38(16):17682--17690.

\bibitem[{Callison-Burch et~al.(2022)Callison-Burch, Tomar, Martin, Ippolito, Bailis, and Reitter}]{callison-burch-etal-2022-dungeons}
Chris Callison-Burch, Gaurav~Singh Tomar, Lara~J. Martin, Daphne Ippolito, Suma Bailis, and David Reitter. 2022.
\newblock \href {https://doi.org/10.18653/v1/2022.emnlp-main.637} {{Dungeons and Dragons as a Dialog Challenge for Artificial Intelligence}}.
\newblock In \emph{Proceedings of the 2022 Conference on Empirical Methods in Natural Language Processing}, pages 9379--9393, Abu Dhabi, United Arab Emirates. Association for Computational Linguistics.

\bibitem[{DeepSeek-AI et~al.(2025)DeepSeek-AI, Guo, Yang, Zhang, Song, Zhang, Xu, Zhu, Ma, Wang, Bi, Zhang, Yu, Wu, Wu, Gou, Shao, Li, Gao, Liu, Xue, Wang, Wu, Feng, Lu, Zhao, Deng, Zhang, Ruan, Dai, Chen, Ji, Li, Lin, Dai, Luo, Hao, Chen, Li, Zhang, Bao, Xu, Wang, Ding, Xin, Gao, Qu, Li, Guo, Li, Wang, Chen, Yuan, Qiu, Li, Cai, Ni, Liang, Chen, Dong, Hu, Gao, Guan, Huang, Yu, Wang, Zhang, Zhao, Wang, Zhang, Xu, Xia, Zhang, Zhang, Tang, Li, Wang, Li, Tian, Huang, Zhang, Wang, Chen, Du, Ge, Zhang, Pan, Wang, Chen, Jin, Chen, Lu, Zhou, Chen, Ye, Wang, Yu, Zhou, Pan, Li, Zhou, Wu, Ye, Yun, Pei, Sun, Wang, Zeng, Zhao, Liu, Liang, Gao, Yu, Zhang, Xiao, An, Liu, Wang, Chen, Nie, Cheng, Liu, Xie, Liu, Yang, Li, Su, Lin, Li, Jin, Shen, Chen, Sun, Wang, Song, Zhou, Wang, Shan, Li, Wang, Wei, Zhang, Xu, Li, Zhao, Sun, Wang, Yu, Zhang, Shi, Xiong, He, Piao, Wang, Tan, Ma, Liu, Guo, Ou, Wang, Gong, Zou, He, Xiong, Luo, You, Liu, Zhou, Zhu, Xu, Huang, Li, Zheng, Zhu, Ma, Tang, Zha, Yan, Ren, Ren, Sha, Fu, Xu, Xie, Zhang,
  Hao, Ma, Yan, Wu, Gu, Zhu, Liu, Li, Xie, Song, Pan, Huang, Xu, Zhang, and Zhang}]{deepseekai2025deepseekr1incentivizingreasoningcapability}
DeepSeek-AI, Daya Guo, Dejian Yang, Haowei Zhang, Junxiao Song, Ruoyu Zhang, Runxin Xu, Qihao Zhu, Shirong Ma, Peiyi Wang, Xiao Bi, Xiaokang Zhang, Xingkai Yu, Yu~Wu, Z.~F. Wu, Zhibin Gou, Zhihong Shao, Zhuoshu Li, Ziyi Gao, and 181 others. 2025.
\newblock \href {https://arxiv.org/abs/2501.12948} {{DeepSeek-R1: Incentivizing Reasoning Capability in LLMs via Reinforcement Learning}}.
\newblock \emph{Preprint}, arXiv:2501.12948.

\bibitem[{Grattafiori et~al.(2024)Grattafiori, Dubey, Jauhri, Pandey, Kadian, Al-Dahle, Letman, Mathur, Schelten, Vaughan, Yang, Fan, Goyal, Hartshorn, Yang, Mitra, Sravankumar, Korenev, Hinsvark, Rao, Zhang, Rodriguez, Gregerson, Spataru, Roziere, Biron, Tang, Chern, Caucheteux, Nayak, Bi, Marra, McConnell, Keller, Touret, Wu, Wong, Ferrer, Nikolaidis, Allonsius, Song, Pintz, Livshits, Wyatt, Esiobu, Choudhary, Mahajan, Garcia-Olano, Perino, Hupkes, Lakomkin, AlBadawy, Lobanova, Dinan, Smith, Radenovic, Guzmán, Zhang, Synnaeve, Lee, Anderson, Thattai, Nail, Mialon, Pang, Cucurell, Nguyen, Korevaar, Xu, Touvron, Zarov, Ibarra, Kloumann, Misra, Evtimov, Zhang, Copet, Lee, Geffert, Vranes, Park, Mahadeokar, Shah, van~der Linde, Billock, Hong, Lee, Fu, Chi, Huang, Liu, Wang, Yu, Bitton, Spisak, Park, Rocca, Johnstun, Saxe, Jia, Alwala, Prasad, Upasani, Plawiak, Li, Heafield, Stone, El-Arini, Iyer, Malik, Chiu, Bhalla, Lakhotia, Rantala-Yeary, van~der Maaten, Chen, Tan, Jenkins, Martin, Madaan, Malo, Blecher,
  Landzaat, de~Oliveira, Muzzi, Pasupuleti, Singh, Paluri, Kardas, Tsimpoukelli, Oldham, Rita, Pavlova, Kambadur, Lewis, Si, Singh, Hassan, Goyal, Torabi, Bashlykov, Bogoychev, Chatterji, Zhang, Duchenne, Çelebi, Alrassy, Zhang, Li, Vasic, Weng, Bhargava, Dubal, Krishnan, Koura, Xu, He, Dong, Srinivasan, Ganapathy, Calderer, Cabral, Stojnic, Raileanu, Maheswari, Girdhar, Patel, Sauvestre, Polidoro, Sumbaly, Taylor, Silva, Hou, Wang, Hosseini, Chennabasappa, Singh, Bell, Kim, Edunov, Nie, Narang, Raparthy, Shen, Wan, Bhosale, Zhang, Vandenhende, Batra, Whitman, Sootla, Collot, Gururangan, Borodinsky, Herman, Fowler, Sheasha, Georgiou, Scialom, Speckbacher, Mihaylov, Xiao, Karn, Goswami, Gupta, Ramanathan, Kerkez, Gonguet, Do, Vogeti, Albiero, Petrovic, Chu, Xiong, Fu, Meers, Martinet, Wang, Wang, Tan, Xia, Xie, Jia, Wang, Goldschlag, Gaur, Babaei, Wen, Song, Zhang, Li, Mao, Coudert, Yan, Chen, Papakipos, Singh, Srivastava, Jain, Kelsey, Shajnfeld, Gangidi, Victoria, Goldstand, Menon, Sharma, Boesenberg,
  Baevski, Feinstein, Kallet, Sangani, Teo, Yunus, Lupu, Alvarado, Caples, Gu, Ho, Poulton, Ryan, Ramchandani, Dong, Franco, Goyal, Saraf, Chowdhury, Gabriel, Bharambe, Eisenman, Yazdan, James, Maurer, Leonhardi, Huang, Loyd, Paola, Paranjape, Liu, Wu, Ni, Hancock, Wasti, Spence, Stojkovic, Gamido, Montalvo, Parker, Burton, Mejia, Liu, Wang, Kim, Zhou, Hu, Chu, Cai, Tindal, Feichtenhofer, Gao, Civin, Beaty, Kreymer, Li, Adkins, Xu, Testuggine, David, Parikh, Liskovich, Foss, Wang, Le, Holland, Dowling, Jamil, Montgomery, Presani, Hahn, Wood, Le, Brinkman, Arcaute, Dunbar, Smothers, Sun, Kreuk, Tian, Kokkinos, Ozgenel, Caggioni, Kanayet, Seide, Florez, Schwarz, Badeer, Swee, Halpern, Herman, Sizov, Guangyi, Zhang, Lakshminarayanan, Inan, Shojanazeri, Zou, Wang, Zha, Habeeb, Rudolph, Suk, Aspegren, Goldman, Zhan, Damlaj, Molybog, Tufanov, Leontiadis, Veliche, Gat, Weissman, Geboski, Kohli, Lam, Asher, Gaya, Marcus, Tang, Chan, Zhen, Reizenstein, Teboul, Zhong, Jin, Yang, Cummings, Carvill, Shepard, McPhie,
  Torres, Ginsburg, Wang, Wu, U, Saxena, Khandelwal, Zand, Matosich, Veeraraghavan, Michelena, Li, Jagadeesh, Huang, Chawla, Huang, Chen, Garg, A, Silva, Bell, Zhang, Guo, Yu, Moshkovich, Wehrstedt, Khabsa, Avalani, Bhatt, Mankus, Hasson, Lennie, Reso, Groshev, Naumov, Lathi, Keneally, Liu, Seltzer, Valko, Restrepo, Patel, Vyatskov, Samvelyan, Clark, Macey, Wang, Hermoso, Metanat, Rastegari, Bansal, Santhanam, Parks, White, Bawa, Singhal, Egebo, Usunier, Mehta, Laptev, Dong, Cheng, Chernoguz, Hart, Salpekar, Kalinli, Kent, Parekh, Saab, Balaji, Rittner, Bontrager, Roux, Dollar, Zvyagina, Ratanchandani, Yuvraj, Liang, Alao, Rodriguez, Ayub, Murthy, Nayani, Mitra, Parthasarathy, Li, Hogan, Battey, Wang, Howes, Rinott, Mehta, Siby, Bondu, Datta, Chugh, Hunt, Dhillon, Sidorov, Pan, Mahajan, Verma, Yamamoto, Ramaswamy, Lindsay, Lindsay, Feng, Lin, Zha, Patil, Shankar, Zhang, Zhang, Wang, Agarwal, Sajuyigbe, Chintala, Max, Chen, Kehoe, Satterfield, Govindaprasad, Gupta, Deng, Cho, Virk, Subramanian, Choudhury,
  Goldman, Remez, Glaser, Best, Koehler, Robinson, Li, Zhang, Matthews, Chou, Shaked, Vontimitta, Ajayi, Montanez, Mohan, Kumar, Mangla, Ionescu, Poenaru, Mihailescu, Ivanov, Li, Wang, Jiang, Bouaziz, Constable, Tang, Wu, Wang, Wu, Gao, Kleinman, Chen, Hu, Jia, Qi, Li, Zhang, Zhang, Adi, Nam, Yu, Wang, Zhao, Hao, Qian, Li, He, Rait, DeVito, Rosnbrick, Wen, Yang, Zhao, and Ma}]{grattafiori2024llama3herdmodels}
Aaron Grattafiori, Abhimanyu Dubey, Abhinav Jauhri, Abhinav Pandey, Abhishek Kadian, Ahmad Al-Dahle, Aiesha Letman, Akhil Mathur, Alan Schelten, Alex Vaughan, Amy Yang, Angela Fan, Anirudh Goyal, Anthony Hartshorn, Aobo Yang, Archi Mitra, Archie Sravankumar, Artem Korenev, Arthur Hinsvark, and 542 others. 2024.
\newblock \href {https://arxiv.org/abs/2407.21783} {{The Llama 3 Herd of Models}}.
\newblock \emph{Preprint}, arXiv:2407.21783.

\bibitem[{Li et~al.(2025)Li, Sedoc, and Sundararajan}]{li2025reasoningtrustingbehaviordeepseek}
Rubing Li, João Sedoc, and Arun Sundararajan. 2025.
\newblock \href {https://arxiv.org/abs/2502.12825} {{Reasoning and the Trusting Behavior of DeepSeek and GPT: An Experiment Revealing Hidden Fault Lines in Large Language Models}}.
\newblock \emph{Preprint}, arXiv:2502.12825.

\bibitem[{Liang et~al.(2024)Liang, Wang, Yan, Ouyang, Hu, and Luo}]{xiaozhan2024roleplayinggamedesign}
Xiaozhan Liang, Yu~Wang, Fengyi Yan, Zehong Ouyang, Yong Hu, and Siyu Luo. 2024.
\newblock \href {https://doi.org/10.1145/3681756.3697949} {Reborn of the white bone demon: Role-playing game design using generative ai in xr}.
\newblock In \emph{SIGGRAPH Asia 2024 Posters}, SA '24, pages 37:1--37:3, Tokyo, Japan. Association for Computing Machinery.

\bibitem[{Lin(2004)}]{rouge}
Chin-Yew Lin. 2004.
\newblock \href {https://aclanthology.org/W04-1013/} {{ROUGE}: A package for automatic evaluation of summaries}.
\newblock In \emph{Text Summarization Branches Out}, pages 74--81, Barcelona, Spain. Association for Computational Linguistics.

\bibitem[{Martin et~al.(2018)Martin, Sood, and Riedl}]{martin2018dungeons}
Lara~J. Martin, Srijan Sood, and Mark Riedl. 2018.
\newblock \href {https://ceur-ws.org/Vol-2321/paper4.pdf} {{Dungeons and DQNs: Toward Reinforcement Learning Agents that Play Tabletop Roleplaying Games}}.
\newblock In \emph{Proceedings of the Joint AIIDE Workshop on Intelligent Narrative Technologies and Workshop on Intelligent Cinematography and Editing (INT-WICED)}, Edmonton, AB, Canada. CEUR-WS.

\bibitem[{Musacchio et~al.(2024)Musacchio, Siciliani, Basile, and Semeraro}]{musacchio-etal-2024-leveraging}
Elio Musacchio, Lucia Siciliani, Pierpaolo Basile, and Giovanni Semeraro. 2024.
\newblock \href {https://aclanthology.org/2024.games-1.7/} {Leveraging large language models for spell-generation in dungeons {\&} dragons}.
\newblock In \emph{Proceedings of the 10th Workshop on Games and Natural Language Processing @ LREC-COLING 2024}, pages 61--69, Torino, Italia. ELRA and ICCL.

\bibitem[{OpenAI et~al.(2024)OpenAI, Jaech, Kalai, Lerer, Richardson, El-Kishky, Low, Helyar, Madry, Beutel, Carney, Iftimie, Karpenko, Passos, Neitz, Prokofiev, Wei, Tam, Bennett, Kumar, Saraiva, Vallone, Duberstein, Kondrich, Mishchenko, Applebaum, Jiang, Nair, Zoph, Ghorbani, Rossen, Sokolowsky, Barak, McGrew, Minaiev, Hao, Baker, Houghton, McKinzie, Eastman, Lugaresi, Bassin, Hudson, Li, de~Bourcy, Voss, Shen, Zhang, Koch, Orsinger, Hesse, Fischer, Chan, Roberts, Kappler, Levy, Selsam, Dohan, Farhi, Mely, Robinson, Tsipras, Li, Oprica, Freeman, Zhang, Wong, Proehl, Cheung, Mitchell, Wallace, Ritter, Mays, Wang, Such, Raso, Leoni, Tsimpourlas, Song, von Lohmann, Sulit, Salmon, Parascandolo, Chabot, Zhao, Brockman, Leclerc, Salman, Bao, Sheng, Andrin, Bagherinezhad, Ren, Lightman, Chung, Kivlichan, O'Connell, Osband, Gilaberte, Akkaya, Kostrikov, Sutskever, Kofman, Pachocki, Lennon, Wei, Harb, Twore, Feng, Yu, Weng, Tang, Yu, Candela, Palermo, Parish, Heidecke, Hallman, Rizzo, Gordon, Uesato, Ward,
  Huizinga, Wang, Chen, Xiao, Singhal, Nguyen, Cobbe, Shi, Wood, Rimbach, Gu-Lemberg, Liu, Lu, Stone, Yu, Ahmad, Yang, Liu, Maksin, Ho, Fedus, Weng, Li, McCallum, Held, Kuhn, Kondraciuk, Kaiser, Metz, Boyd, Trebacz, Joglekar, Chen, Tintor, Meyer, Jones, Kaufer, Schwarzer, Shah, Yatbaz, Guan, Xu, Yan, Glaese, Chen, Lampe, Malek, Wang, Fradin, McClay, Pavlov, Wang, Wang, Murati, Bavarian, Rohaninejad, McAleese, Chowdhury, Chowdhury, Ryder, Tezak, Brown, Nachum, Boiko, Murk, Watkins, Chao, Ashbourne, Izmailov, Zhokhov, Dias, Arora, Lin, Lopes, Gaon, Miyara, Leike, Hwang, Garg, Brown, James, Shu, Cheu, Greene, Jain, Altman, Toizer, Toyer, Miserendino, Agarwal, Hernandez, Baker, McKinney, Yan, Zhao, Hu, Santurkar, Chaudhuri, Zhang, Fu, Papay, Lin, Balaji, Sanjeev, Sidor, Broda, Clark, Wang, Gordon, Sanders, Patwardhan, Sottiaux, Degry, Dimson, Zheng, Garipov, Stasi, Bansal, Creech, Peterson, Eloundou, Qi, Kosaraju, Monaco, Pong, Fomenko, Zheng, Zhou, McCabe, Zaremba, Dubois, Lu, Chen, Cha, Bai, He, Zhang, Wang,
  Shao, and Li}]{jaech2024openai}
OpenAI, Aaron Jaech, Adam Kalai, Adam Lerer, Adam Richardson, Ahmed El-Kishky, Aiden Low, Alec Helyar, Aleksander Madry, Alex Beutel, Alex Carney, Alex Iftimie, Alex Karpenko, Alex~Tachard Passos, Alexander Neitz, Alexander Prokofiev, Alexander Wei, Allison Tam, Ally Bennett, and 243 others. 2024.
\newblock \href {https://arxiv.org/abs/2412.16720} {{OpenAI o1 System Card}}.
\newblock \emph{Preprint}, arXiv:2412.16720.

\bibitem[{Papazov et~al.(2022)Papazov, Gill, García~Ferreiro, Zhu, Martin, and Callison-Burch}]{papazov2022avrae}
Stefan Papazov, Wesley Gill, Marta García~Ferreiro, Andrew Zhu, Lara~J. Martin, and Chris Callison-Burch. 2022.
\newblock \href {https://openreview.net/forum?id=jQSStHwtmDN} {{Using Language Models to Convert Between Natural Language and Game Commands}}.
\newblock In \emph{{Wordplay: Where Language Meets Games Workshop at NAACL 2022}}, Seattle, WA.

\bibitem[{Papineni et~al.(2002)Papineni, Roukos, Ward, and Zhu}]{bleu}
Kishore Papineni, Salim Roukos, Todd Ward, and Wei-Jing Zhu. 2002.
\newblock \href {https://doi.org/10.3115/1073083.1073135} {{BLEU}: a method for automatic evaluation of machine translation}.
\newblock In \emph{Proceedings of the 40th Annual Meeting of the Association for Computational Linguistics}, pages 311--318, Philadelphia, Pennsylvania, USA. Association for Computational Linguistics.

\bibitem[{Rae et~al.(2021)Rae, Borgeaud, Cai, Millican, Hoffmann, Song, Aslanides, Henderson, Ring, Young et~al.}]{rae2021scaling}
Jack~W Rae, Sebastian Borgeaud, Trevor Cai, Katie Millican, Jordan Hoffmann, Francis Song, John Aslanides, Sarah Henderson, Roman Ring, Susannah Young, and 1 others. 2021.
\newblock \href {https://arxiv.org/abs/2112.11446} {{Scaling Language Models: Methods, Analysis \& Insights from Training Gopher}}.
\newblock \emph{Preprint}, arXiv:2112.11446.

\bibitem[{Shao et~al.(2024)Shao, Wang, Zhu, Xu, Song, Bi, Zhang, Zhang, Li, Wu, and Guo}]{shao2024deepseekmathpushinglimitsmathematical}
Zhihong Shao, Peiyi Wang, Qihao Zhu, Runxin Xu, Junxiao Song, Xiao Bi, Haowei Zhang, Mingchuan Zhang, Y.~K. Li, Y.~Wu, and Daya Guo. 2024.
\newblock \href {https://arxiv.org/abs/2402.03300} {{DeepSeekMath: Pushing the Limits of Mathematical Reasoning in Open Language Models}}.
\newblock \emph{Preprint}, arXiv:2402.03300.

\bibitem[{Shojaee et~al.(2025)Shojaee, Mirzadeh, Alizadeh, Horton, Bengio, and Farajtabar}]{illusion-of-thinking}
Parshin Shojaee, Iman Mirzadeh, Keivan Alizadeh, Maxwell Horton, Samy Bengio, and Mehrdad Farajtabar. 2025.
\newblock \href {https://arxiv.org/abs/2506.06941} {{The Illusion of Thinking: Understanding the Strengths and Limitations of Reasoning Models via the Lens of Problem Complexity}}.
\newblock \emph{Preprint}, arXiv:2506.06941.

\bibitem[{Shuster et~al.(2022)Shuster, Komeili, Adolphs, Roller, Szlam, and Weston}]{shuster-etal-2022-language}
Kurt Shuster, Mojtaba Komeili, Leonard Adolphs, Stephen Roller, Arthur Szlam, and Jason Weston. 2022.
\newblock \href {https://doi.org/10.18653/v1/2022.findings-emnlp.27} {Language models that seek for knowledge: Modular search {\&} generation for dialogue and prompt completion}.
\newblock In \emph{Findings of the Association for Computational Linguistics: EMNLP 2022}, pages 373--393, Abu Dhabi, United Arab Emirates. Association for Computational Linguistics.

\bibitem[{Song et~al.(2024)Song, Zhu, and Callison-Burch}]{song2024thirteenhourssolvelabyrinth}
Jaewoo Song, Andrew Zhu, and Chris Callison-Burch. 2024.
\newblock \href {https://wordplay-workshop.github.io/wordplay2024/pdfs/11.pdf} {{You Have Thirteen Hours in Which to Solve the Labyrinth: Enhancing AI Game Masters with Function Calling}}.
\newblock In \emph{Wordplay: When Language Meets Games Workshop (ACL 2024)}, Bangkok, Thailand.

\bibitem[{Urbanek et~al.(2019)Urbanek, Fan, Karamcheti, Jain, Humeau, Dinan, Rockt{\"a}schel, Kiela, Szlam, and Weston}]{urbanek2019learning}
Jack Urbanek, Angela Fan, Siddharth Karamcheti, Saachi Jain, Samuel Humeau, Emily Dinan, Tim Rockt{\"a}schel, Douwe Kiela, Arthur Szlam, and Jason Weston. 2019.
\newblock \href {https://doi.org/10.18653/v1/D19-1062} {Learning to speak and act in a fantasy text adventure game}.
\newblock In \emph{Proceedings of the 2019 Conference of the North American Chapter of the Association for Computational Linguistics (NAACL)}, page 673–683.

\bibitem[{Wang et~al.(2023)Wang, Kordi, Mishra, Liu, Smith, Khashabi, and Hajishirzi}]{wang-etal-2023-self-instruct}
Yizhong Wang, Yeganeh Kordi, Swaroop Mishra, Alisa Liu, Noah~A. Smith, Daniel Khashabi, and Hannaneh Hajishirzi. 2023.
\newblock \href {https://doi.org/10.18653/v1/2023.acl-long.754} {Self-instruct: Aligning language models with self-generated instructions}.
\newblock In \emph{Proceedings of the 61st Annual Meeting of the Association for Computational Linguistics (Volume 1: Long Papers)}, pages 13484--13508, Toronto, Canada. Association for Computational Linguistics.

\bibitem[{Wei et~al.(2022)Wei, Wang, Schuurmans, Bosma, ichter, Xia, Chi, Le, and Zhou}]{wei2022chain}
Jason Wei, Xuezhi Wang, Dale Schuurmans, Maarten Bosma, brian ichter, Fei Xia, Ed~Chi, Quoc~V Le, and Denny Zhou. 2022.
\newblock \href {https://proceedings.neurips.cc/paper_files/paper/2022/file/9d5609613524ecf4f15af0f7b31abca4-Paper-Conference.pdf} {Chain-of-thought prompting elicits reasoning in large language models}.
\newblock In \emph{Advances in Neural Information Processing Systems}, volume~35, pages 24824--24837. Curran Associates, Inc.

\bibitem[{Yao et~al.(2023)Yao, Yu, Zhao, Shafran, Griffiths, Cao, and Narasimhan}]{yao2023tree}
Shunyu Yao, Dian Yu, Jeffrey Zhao, Izhak Shafran, Thomas~L. Griffiths, Yuan Cao, and Karthik~R Narasimhan. 2023.
\newblock \href {https://proceedings.neurips.cc/paper_files/paper/2023/hash/271db9922b8d1f4dd7aaef84ed5ac703-Abstract-Conference.html} {{Tree of Thoughts: Deliberate Problem Solving with Large Language Models}}.
\newblock In \emph{Thirty-seventh Conference on Neural Information Processing Systems}, volume~36, pages 11809--11822. Curran Associates, Inc.

\bibitem[{Yin et~al.(2025)Yin, Jiang, Chen, Wang, and Ling}]{yin2025enhancinggeneralizationchainthought}
Maxwell~J. Yin, Dingyi Jiang, Yongbing Chen, Boyu Wang, and Charles Ling. 2025.
\newblock \href {https://arxiv.org/abs/2501.09804} {Enhancing generalization in chain of thought reasoning for smaller models}.
\newblock \emph{Preprint}, arXiv:2501.09804.

\bibitem[{Zhou et~al.(2023)Zhou, Sch{\"a}rli, Hou, Wei, Scales, Wang, Schuurmans, Cui, Bousquet, Le, and Chi}]{zhou2023leasttomost}
Denny Zhou, Nathanael Sch{\"a}rli, Le~Hou, Jason Wei, Nathan Scales, Xuezhi Wang, Dale Schuurmans, Claire Cui, Olivier Bousquet, Quoc~V Le, and Ed~H. Chi. 2023.
\newblock \href {https://openreview.net/forum?id=WZH7099tgfM} {Least-to-most prompting enables complex reasoning in large language models}.
\newblock In \emph{The Eleventh International Conference on Learning Representations}, Kigali, Rwanda.

\bibitem[{Zhu et~al.(2023)Zhu, Aggarwal, Feng, Martin, and Callison-Burch}]{Zhu_2023}
Andrew Zhu, Karmanya Aggarwal, Alexander Feng, Lara Martin, and Chris Callison-Burch. 2023.
\newblock \href {https://doi.org/10.18653/v1/2023.acl-long.229} {{FIREBALL: A Dataset of Dungeons and Dragons Actual-Play with Structured Game State Information}}.
\newblock In \emph{Proceedings of the 61st Annual Meeting of the Association for Computational Linguistics (Volume 1: Long Papers)}, page 4171–4193. Association for Computational Linguistics.

\end{thebibliography}

\appendix 
\section{Appendix}
\subsection{Compute Details}
\label{sec:app-compute}
The compute we used for the experiments for this paper used a data science machine with two NVIDIA GPU A6000s with 48 GB of GPU memory connected with NVLINK. One model was run at a time to batch inference a dataset of 20355 input prompts that contained 4071 distinct game state samples appended to five different input prompts.   The duration of the batch run for LLaMA-3.1-8B-Instruct was 3:10:28 hours.  Deepseek-R1-Distill-LLaMA-8B took just over 8:56:28 hours.

\subsection{Full Example - combat\_state\_before}

\begin{tcolorbox}[breakable,colback=red!3!white,colframe=red!75!black,title=combat\_state\_before]
\texttt{{\textcolor{red}{"name"}: "Emma Thornwall", \\ 
\textcolor{red}{"hp"}: "<80/80 HP; Healthy>", \\
\textcolor{red}{"class"}: "Warlock 11", "race": "Eladrin", \\ 
\textcolor{red}{"attacks"}: "Crossbow, light, Dagger, Quarterstaff, 2-Handed Quarterstaff, Scross, Silver, Unarmed Strike", \\ 
\textcolor{red}{"spells"}: "Levitate, Magic Missile, Hold Person, Counterspell, Witch Bolt, Intellect Fortress, Dimension Door, Raulothim's Psychic Lance, Sending, Death Ward, Polymorph, Hex, Mirror Image, Eyebite, Eldritch Blast, Prestidigitation, Mage Hand, Dissonant Whispers, Mind Sliver", \\ 
\textcolor{red}{"actions"}: "Sculptor of Flesh, Fey Step (Winter), Protection of the Talisman, Fey Step (Summer), Fey Step (Spring), Agonizing Blast, Pact of the Talisman, Fey Step (Autumn), Entropic Ward, Ascendant Step", \\ 
\textcolor{red}{"effects"}: "Blessed", \\ 
\textcolor{red}{"description"}: null, \\ 
\textcolor{red}{"controller\_id"}: "178818952437053304"}, 
{\textcolor{red}{"name"}: "BA1", \\ 
\textcolor{red}{"hp"}: "<262/262 HP; Healthy>", \\ 
\textcolor{red}{"class"}: null, \\ 
\textcolor{red}{"race"}: "Balor", \\ 
\textcolor{red}{"attacks"}: "Death Throes, Fire Aura, Longsword, Whip", \\ 
\textcolor{red}{"spells"}: "", \\ 
\textcolor{red}{"actions"}: null, \\
\textcolor{red}{"effects"}: "", \\
\textcolor{red}{"description"}: null, \\
\textcolor{red}{"controller\_id"}: "159332117133051198", \\ 
\textcolor{red}{"name"}: "Jaguar", \\ 
\textcolor{red}{"hp"}: "<91/91 HP; Healthy>", \\ 
\textcolor{red}{"class"}: "Barbarian 7/Cleric 2", \\
\textcolor{red}{"race"}: "Goliath", \\ 
\textcolor{red}{"attacks"}: "Greatsword, +1, Javelin, Javelin of Lightning, Javelin of Lightning2, Javelin of Lightning3, Javelin of Lightning4, Javelin of Lightning5, Javelin of Lightning6, Unarmed Strike", \\ 
\textcolor{red}{"spells"}: "Sacred Flame, Divine Favor, Cure Wounds, Fireball, Bless, Ceremony, Shield of Faith, Guiding Bolt, Detect Evil and Good, Spare the Dying, Speak with Animals, Guidance", \\ 
\textcolor{red}{"actions"}: "Stone's Endurance, Channel Divinity: Guided Strike (War Domain), Channel Divinity: Turn Undead, Rage, War Priest, Bear Rage, Channel Divinity",\\ 
\textcolor{red}{"effects"}: "Bless, Blessed", \\ 
\textcolor{red}{"description"}: null, \\ 
\textcolor{red}{"controller\_id"}: "743069156064144035"}, 
{\textcolor{red}{"name"}: "Zenthaea", \\ 
\textcolor{red}{"hp"}: "<92/92 HP; Healthy>", \\ 
\textcolor{red}{"class"}: "Fighter 11", "race": "Eladrin", \\ 
\textcolor{red}{"attacks"}: "Crossbow, light, Stardust, 2-Handed Stardust, Unarmed Strike", \\ \textcolor{red}{"spells"}: "Blur, Jim's Magic Missile, Booming Blade, Hold Person, Mirror Image, Shield, Absorb Elements, Chromatic Orb, Enlarge/Reduce, Blade Ward, Fire Bolt, Death Ward", \\ \textcolor{red}{"actions"}: "Petal Cloud (Autumn), Weapon Bond, Petal Cloud (Winter), Indomitable, Shield Master Shove, Shield Master Evasion, War Magic, Petal Cloud (Summer), Maneuvers: Riposte, Maneuvers: Goading Attack, Extra Attack, Feywild's Gift, Martial Adept, Stardust's Zeal, Petal Cloud (Spring)", \\ 
\textcolor{red}{"effects"}: "Enlarge/Reduce, Enlarged/Reduced", \\ 
\textcolor{red}{"description"}: "A young, pale blue eladrin who carries an opalescent sword and a large shield, each in one hand.", \\ 
\textcolor{red}{"controller\_id"}: "233849403576355489"}, 
{\textcolor{red}{"name"}: "Petcan Gard", \\ 
\textcolor{red}{"hp"}: "<80/80 HP; Healthy>", \\ 
\textcolor{red}{"class"}: "Cleric 11", "race": "Eladrin", \\ 
\textcolor{red}{"attacks"}: "Hellfire Mace, Unarmed Strike", \\ 
\textcolor{red}{"spells"}: "Lesser Restoration, Resistance, Inflict Wounds, Hold Person, Greater Restoration, Guiding Bolt, Healing Word, Spare the Dying, Spiritual Weapon, Banishment, Death Ward, Sacred Flame, Toll the Dead, Blindness/Deafness, Dispel Magic, Beacon of Hope, Animate Dead, Spirit Guardians, Bless, Mass Cure Wounds, Mass Healing Word, Revivify, Guardian of Faith, Cure Wounds, Heal, Spirit Shroud, Shield of Faith, Guidance, Raise Dead", \\ 
\textcolor{red}{"actions"}: "Divine Intervention, Blessed Healer, Channel Divinity: Turn Undead, Healer Healing, Fey Step (Summer), Fey Step (Spring), Disciple of Life, Channel Divinity: Preserve Life, Fey Step (Autumn), Fey Step (Winter), Channel Divinity", \\ 
\textcolor{red}{"effects"}: "Blessed", \\ 
\textcolor{red}{"description"}: null, \\ 
\textcolor{red}{"controller\_id"}: "227155263075781005"}}
\end{tcolorbox}

\end{document}